\documentclass[10pt,twocolumn,letterpaper]{article}

\usepackage{iccv}
\usepackage{times}
\usepackage{epsfig}
\usepackage{graphicx}
\usepackage{amsmath}
\usepackage{amssymb}
\usepackage{booktabs}
\usepackage{multirow}
\usepackage{multicol}
\usepackage{cite}
\usepackage{comment}
\usepackage[export]{adjustbox}

\usepackage[pagebackref=true,breaklinks=true,letterpaper=true,colorlinks,bookmarks=false]{hyperref}

\iccvfinalcopy 

\newcommand{\reals}{\mathbb{R}}

\usepackage[capitalize]{cleveref}
\crefname{section}{Sec.}{Secs.}
\Crefname{section}{Section}{Sections}
\Crefname{table}{Table}{Tables}
\crefname{table}{Tab.}{Tabs.}

\def\iccvPaperID{6373} 

\ificcvfinal\fi

\begin{document}


\title{NeRFMeshing: Distilling Neural Radiance Fields\\into Geometrically-Accurate 3D Meshes}




\author{
Marie-Julie Rakotosaona~\textsuperscript{1} \quad
Fabian Manhardt~\textsuperscript{1} \quad
Diego Martin Arroyo~\textsuperscript{1} \quad \\
Michael Niemeyer~\textsuperscript{1} \quad
Abhijit Kundu~\textsuperscript{1} \quad
Federico Tombari~\textsuperscript{1,2} \quad
\\[1em]
\textsuperscript{1} Google
\quad
\textsuperscript{2}{TU Munich}
\vspace*{-5mm}
}

\maketitle
\ificcvfinal\thispagestyle{empty}\fi

\begin{abstract}
With the introduction of Neural Radiance Fields (NeRFs), novel view synthesis has recently made a big leap forward. 
At the core, NeRF proposes that each 3D point can emit radiance, allowing to conduct view synthesis using differentiable volumetric rendering. While neural radiance fields can accurately represent 3D scenes for computing the image rendering, 3D meshes are still the main scene representation supported by most computer graphics and simulation pipelines, enabling tasks such as real time rendering and physics-based simulations.  
Obtaining 3D meshes from neural radiance fields still remains an open challenge since NeRFs are optimized for view synthesis, not enforcing an accurate underlying geometry on the radiance field.
We thus propose a novel compact and flexible architecture that enables easy 3D surface reconstruction from any NeRF-driven approach.
Upon having trained the radiance field, we distill the volumetric 3D representation into a Signed Surface Approximation Network, allowing easy extraction of the 3D mesh and appearance.
Our final 3D mesh is physically accurate and can be rendered in real time on an array of devices.
\end{abstract}


\section{Introduction}

\begin{figure}
  \centering
   \includegraphics[width=1.0\linewidth]{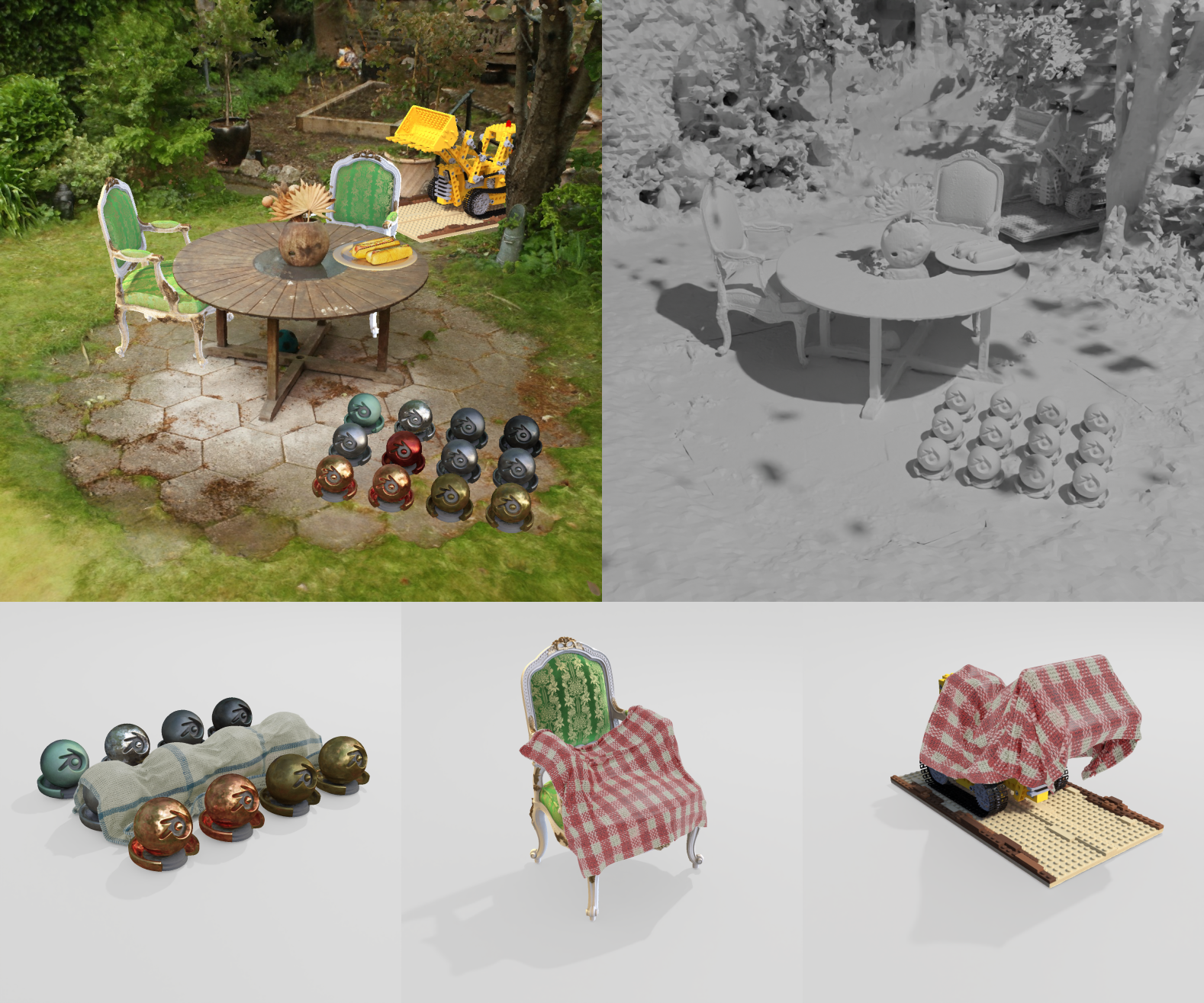}
   \caption{Our method extracts meshes with accurate geometry and view dependent appearance given a collection of posed images.  We show a composition of meshes extracted from the \textit{Chair}, \textit{Hotdog}, \textit{Lego} (Blender dataset) and \textit{Garden} (MipNeRF 360 dataset) scenes using our method (Top left: scene rendered with colors, top right: geometry visualization). Our method enables physics based simulations. We show the results of a simulation of a cloth falling on the objects from the Blender Synthetic dataset  (Bottom). }
   \label{fig:teaser}
\end{figure}

Accurate 3D scene and object reconstruction is a key problem in areas such as robotics, photogrammetry, AR/VR, where applications often rely on precise 3D geometry to perform physics-based simulations, real-time 3D visualizations, rendering and interactions. 
Moreover, the related field of novel view synthesis (NVS) has made tremendous advances in recent years.
Recently Mildenhall \textit{et al.} \cite{mildenhall2020nerf} proposed to perform NVS by means of neural radiance fields (NeRFs), a novel 3D representation where each 3D location in space can emit radiance, see Sec \ref{sec-nerfs} for more details. Novel views are  synthesized by means of differentiable volumetric rendering~\cite{mildenhall2020nerf,xie2022neural}. Due to the impressive results and simplicity of the approach, most related work has focused on improving NeRF in terms of image quality~\cite{barron2022mipnerf360}, robustness~\cite{lin2021barf,niemeyer2022regnerf,martin2021nerf}, as well as training speed~\cite{mueller2022instant} and rendering speed~\cite{hedman2021baking,chen2022mobilenerf}. 
Unfortunately, as these representations are commonly optimized for the NVS task and not explicitly for the underlying geometry \cite{chen2022mobilenerf, hedman2021snerg}, it is yet unclear how to best obtain accurate 3D meshes from radiance fields.  Indeed, while the volumetric representation of NeRF enables accurate renderings from new views, the underlying 3D geometry of each object is not uniquely defined as a level-set surface. NeRF methods often rely on layering and transparency effects to approximate complex appearance and geometry. The surface of objects is therefore approximated by dense regions of the volume instead of surfaces of zero thickness. Moreover, most related work still lacks the capability to be rendered in real-time \cite{verbin2022ref, niemeyer2022regnerf}, especially on commodity hardware. Finally, NeRFs cannot be directly integrated with most computer graphics (CG) pipelines, as they still rely on standard 3D meshes due to their compactness and physical properties.

Despite some recent work proposing alternative scene representations to re-enable real-time rendering even for NeRFs, they are again not designed to produce accurate 3D representations of the input objects or scenes to be used with standard CG pipelines~\cite{hedman2021snerg, chen2022mobilenerf}.

To deal with these limitations, we thus introduce \emph{NeRFMeshing}, an end-to-end pipeline for efficiently extracting geometrically accurate meshes from trained NeRF-based networks, merely adding a very small overhead in time. Our method produces meshes with neural colors having accurate geometry that can be rendered in real time on commodity hardware. 
Introducing a novel signed surface approximation network (SSAN), we train a post-processing NeRF pipeline, defining the underlying surface and appearance. SSAN produces an accurate 3D triangle mesh of the scene that we render using a small appearance network to produce view-dependent colors. In contrast to other works that leverage distance fields, requiring significant modifications in the used NeRF architecture~\cite{yariv2021volume, wang2021neus}, our method can be leveraged together with any NeRF, enabling to easily incorporate new advances, such as improved handling of unbounded scenes~\cite{barron2022mipnerf360} or reflective objects~\cite{verbin2022ref}. 
Essentially, SSAN estimates a Truncated Signed Distance Field (TSDF) and a feature appearance field. Harnessing the NeRF approximated geometry as well as the used training views, we distill the trained NeRF into the SSAN model. 
We then extract the 3D mesh from the trained SSAN which can be rendered on embedded devices at high frame-rate using rasterization and the appearance network.  Thanks to the flexibility of our method, we can generate these 3D meshes fast ~\cite{mueller2022instant}, are not tied to object-centric scenes~\cite{barron2022mipnerf360}, and can even model complex and non-lambertian surfaces~\cite{verbin2022ref}. 

To summarize, we propose NeRFMeshing, a novel method for capturing both accurate 3D meshes of the scene as well as enabling realistic view dependent rendering. The extracted meshes from our end-to-end pipeline can be integrated in graphics and simulation pipelines. Our model also preserves the high fidelity of neural radiance fields like view-dependent effects and reflections and can be used for real-time novel view synthesis.

\section{Related Work}
The base of our work is the neural radiance field (NeRF) formulation initially introduced in~\cite{mildenhall2020nerf}. This work describes a trainable radiance field parameterized with a neural network. Subsequent works have addressed the main limitations of the original approach, such as the slow training speed~\cite{mueller2022instant}, anti-aliasing effects or the ability to model unbounded scenes~\cite{barron2022mipnerf360}. Another main caveat of the original formulation is the lack of accurate underlying geometry, mostly caused by the fact that NeRFs are optimized exclusively for visual consistency. This gives way for the network to create occupancy regions to support the volumetric rendering process even when particular parts of the space are not occupied by an underlying surface. Additional supervision signals, such as depth~\cite{kangle2021dsnerf, roessle2022depthpriorsnerf} are effective in regularizing the geometry, but require additional input modalities which might be difficult to source. In contrast, our work relies on NeRF networks trained from images.

An alternative to radiance fields is to learn a Signed Distance Function (SDF)~\cite{wang2021neus, oechsle2021unisurf, yariv2021volume}. A high-quality mesh can be extracted by means of marching cubes or similar approaches.  \cite{munkberg2021nvdiffrec} combines this approach with a tetrahedral grid that is optimized during training, followed by differentiable rasterizer to recover materials and lighting. However, this method is constrained by the choice of the grid resolution at training time. Compared to \cite{munkberg2021nvdiffrec}, our method does not rely on a fixed grid template during training, enabling us to reconstruct at any arbitrary resolution afterwards.  Similarly, \cite{yariv2020multiview} uses an SDF with sphere tracing to determine the intersection with the surface.  In this work, we are mainly interested in obtaining a unique and geometrically accurate 3D mesh surface from NeRF methods. We exploit the adaptive power of NeRFs to be able to robustly represent 3D scenes in a range of conditions and environments. In particular, we do not require any changes to the NeRF architecture in order to compute a surface. Also differentiable mesh rasterizers often suffer from optimization issues which are easily avoidable in differentiable volumetric rendering adopted by the NeRF methods.

Recent approaches advance speed and geometric accuracy of NeRFs. SNeRG~\cite{hedman2021snerg} has achieved real-time rendering of radiance fields by restructuring the original NeRF architecture to precompute the predicted density, view-independent colors and feature vectors in a sparse voxel grid. A separate  view-dependent network runs online to compute the final color. It results in a real-time visualization with a good trade-off in image quality, however, the underlying geometry is not well defined. MobileNeRF~\cite{chen2022mobilenerf} builds on a similar principle, but relies on a triangle mesh that is optimized during training as the underlying geometry. However, the final geometry is far from being accurate. As the method relies on a single resolution grid at training time, the faces from the triangle soup cannot fit detailed regions. Moreover, multiple layers of faces can sometimes describe the same surface during optimization leading to inaccurate geometry to support the rendering loss. Voxel grids are also a viable alternative to efficiently store geometry ~\cite{yu2021plenoctrees,yu_and_fridovichkeil2021plenoxels} and have efficient renderings. However, they are inherently constrained by the grid resolution and can suffer from discretization artifacts.

\section{Method}

\begin{figure*}
  \centering
  \includegraphics[width=1.\linewidth]{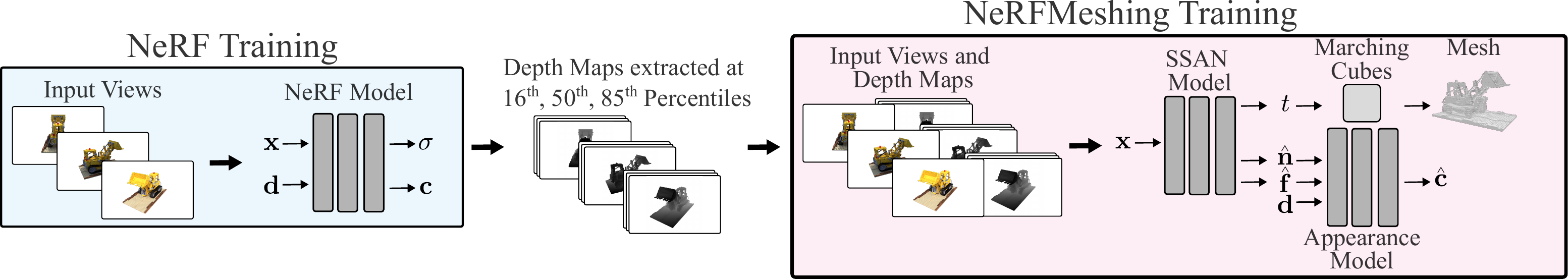}
     \caption{We exploit rendered depth distribution from NeRF to help supervise an approximated TSDF. We produce learned features that we feed to a small appearance network to predict RGB colors. We can extract the surface using marching cubes and store appearance features on the surface. Finally we render in real time using the mesh, appearance features and appearance network.}
   \label{fig:pipeline}
\end{figure*}

In this section we present our approach for extracting accurate 3D meshes with neural features from NeRF for subsequent real-time rendering. We present an overview of the method in \cref{fig:pipeline}. We first briefly outline the general concept of NeRFs in \cref{sec-nerfs}. In \cref{sec-surface-approx} we present our method for approximating surface from NeRF. Finally in \cref{sec-rendering} we describe mesh extraction and real-time rendering.

\subsection{Neural Radiance Fields} \label{sec-nerfs}
At the core, a neural radiance field is a continuous mapping from a 3D location $\mathbf{x} \in \mathbb{R}^3$ and a ray viewing direction

 $\mathbf{d} \in \mathbb{S}^2$ 
to an RGB color $\mathbf{c} \in \left[ 0, 1 \right]^3$ and volume density $\sigma \in \reals^+$. It can be formulated as,
\begin{equation}
\left[\mathbf{c}, \sigma\right] = F_{\theta} \left(\gamma_{\textit{x}}(\mathbf{x}), \gamma_{\textit{d}}({\mathbf{d}})\right) \,
\label{eq:mlp}
\end{equation}
where $F$ is modeled as an MLP with learnable parameters  $\theta$, and $\gamma: \reals^{3} \rightarrow \reals^{N}$ is a positional encoding of the input $\mathbf{x}$ required to capture high frequencies.

Given a camera pose $P_i = [R_i, t_i] \in SE(3)$, each pixel coordinate $\mathbf{p} \in \reals^{2}$  determines a ray in the world coordinate system, whose origin is the camera center of projection $\mathbf{o}_{i} = \mathbf{t}_i$ and direction is defined as $\mathbf{d}_{i, \text{p}} = R_i K_i^{-1}\bar{\mathbf{p}}$. We can express a 3D point along the viewing ray associated with $\mathbf{p}$ at depth $t$ as $\mathbf{r}_{i, \text{p}} (t) = \mathbf{o}_i + t \mathbf{d}_{i, \text{p}}$. 
To render the color $\hat{\mathbf{I}}_{i, \text{p}} \in \left[0, 1\right]^3$ at pixel $\mathbf{p}$, we sample $M$ discrete depth values $t_m$ along the ray within the near and far plane $\left[t_n, t_f \right]$, and query $F_{\theta}$ at the associated 3D points. The corresponding predicted color and volume density values $\left\{(\mathbf{c}_m, \sigma_m  ) \right\}_{m=1}^M$ are then composited as,
\begin{align}
\hat{\mathbf{I}}_{i, \text{p}} &= \hat{I}(\mathbf{p}; \theta, P_i) = \sum_{m=1}^{M} \alpha_m\mathbf{c}_m \,, \label{eq:volume_rendering} \\
\text{where} \quad \alpha_m &= T_m\left(1 - \exp(-\sigma_m\delta_m)\right) \,, \label{eq:rendering_weight} \\
T_m &= \exp \left( -\sum_{m'=1}^{m} \sigma_{m'}\delta_{m'}\right)\,. \label{eq:transmittance}
\end{align}

where $T_m$ denotes the accumulated transmittance along the ray from $t_n$ to $t_m$, and $\delta_m = t_{m+1} - t_m$ is the distance between adjacent samples. 
Finally we compute accumulated depth at percentile $k$ as follows,
\begin{equation}\label{eq:rendered-depth}
    \hat{z}_{i, \text{p}, k}  =  \sum_{m=1}^{M_k} \alpha_m t_m \,. 
\end{equation}
Where $M_k$ denotes the first index at which the accumulated transmittance $T_k > k/100$. 

\subsection{Surface approximation from NeRF} \label{sec-surface-approx}

In this section we introduce our signed surface approximation network (SSAN) module that creates a truncated signed distance field (TSDF) from NeRFs.

\paragraph{Signed Surface Approximation Network (SSAN). } SSAN is optimized from pretrained NeRFs to enable the extraction of a unique and accurate surface, and appearance. Relying on the pre-trained NeRF representations brings two main advantages, which we harness when training the SSAN: a rough 3D approximation of the scene geometry learnt by the NeRF, and useful priors such as rendered depths via volumetric rendering. 
To this end, we feed a 3D coordinate $\mathbf{x} \in \mathbb{R}^{3}$ to SSAN $\phi : \mathbb{R}^{3} \rightarrow (\mathbb{R}, \mathbb{S}^{2}, \mathbb{R}^{8})$ to predict the truncated signed distance approximation $\hat{t} \in \left[ -0.1, 0.1 \right]$, the normal $\mathbf{\hat{n}} \in \mathbb{S}^2$, used for normal smoothness regularization as well as rendering, and 8-dimensional appearance feature $\mathbf{\hat{f}} \in \left[ 0, 1 \right]^{8}$ with 
\begin{equation}
    [\hat{t}, \mathbf{\hat{n}}, \mathbf{\hat{f}}] = \phi(\mathbf{x}).
\end{equation} The outgoing features $\mathbf{\hat{f}}$ and normals $\mathbf{\hat{n}}$ are then further processed together with the viewing direction $\mathbf{d} \in \mathbb{S}^{2}$ by a small appearance network $\eta: (\mathbb{R}^8, \mathbb{S}^2, \mathbb{S}^2) \rightarrow \mathbb{R}^{3}$ to produce the view-dependant RGB color $\mathbf{\hat{c}} \in \left[ 0, 1 \right]^3$ according to
\begin{equation}
    \mathbf{\hat{c}} = \eta(\mathbf{\hat{f}}, \mathbf{\hat{n}}, \mathbf{d}).
\end{equation} 
We refer to Fig.~\ref{fig:pipeline} for an abstract overview over of SSAN.

\paragraph{TSDF.}  We aim to learn a TSDF approximation $t$ that represents the underlying surface of a scene or object. $t$ should have an accurate sign globally (positive outside the surface and negative inside).  Moreover,  $t$ should be smooth and have near constant derivatives close to the surface to enable efficient marching cubes at a later stage. Note that  we are not constrained to produce a well defined distance function. In this section, we introduce multiple loss functions to enforce these properties.

 At training time, we exploit the NeRF occupancy aggregated along rays. While the weighted average of the occupancy is often used to render an approximation of depth, we instead use the distribution of the occupancy along the ray for more accuracy as in~\cite{park2021nerfies}. We call $\hat{z}_{i, \text{p}, 50}$ the median value of depths along a ray as described in \cref{sec-nerfs}. We can estimate 3D points that are outside or inside the surface by constructing points before or after the median depth from the camera origin respectively. In practice, we choose to render  the depth of the NeRF outside at the $16^{th}$ percentile $\hat{z}_{i, \text{p}, 16}$ and inside the object surface at the $84^{th}$ percentile $\hat{z}_{i, \text{p}, 84}$. Subsequently, we compute the corresponding 3D world coordinates and feed them to SSAN to estimate the respective SDFs with:

\begin{align}
    \left(\mathbf{\hat{t}}_{50, i, p}, \mathbf{\hat{n}}_{50, i}, \mathbf{\hat{f}}_{50, i}\right)  & =  \phi(r_{i,p}(\hat{z}_{i, \text{p}, 50})), \\ 
    \left(\mathbf{\hat{t}}_{16, i, p}, \mathbf{\hat{n}}_{16, i}, \mathbf{\hat{f}}_{16, i}\right)  & =  \phi(r_{i,p}(\hat{z}_{i, \text{p}, 16})), \quad \text{and} \\ 
    \left(\mathbf{\hat{t}}_{84, i, p}, \mathbf{\hat{n}}_{84, i}, \mathbf{\hat{f}}_{84, i}\right) & = \phi(r_{i,p}(\hat{z}_{i, \text{p}, 84})).
\end{align} 

For readability, we drop the term $p$ in later references. During training, we enforce this construction by applying the loss $L_i$  at sampled projections along training rays.  

\begin{figure}
  \centering
   \includegraphics[width=0.7\linewidth]{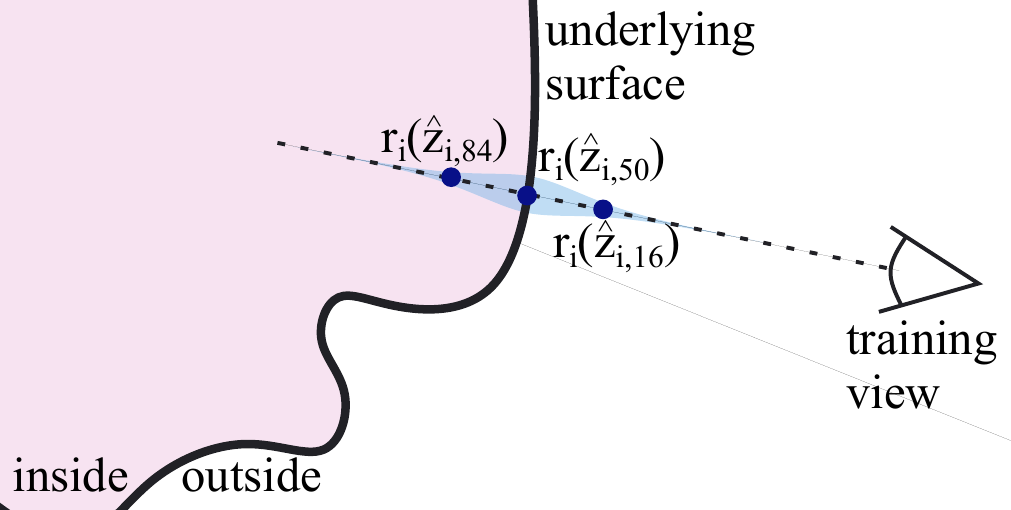}
   \caption{We exploit occupancy aggregated over training rays to estimate the outside ($ r_{i}(\hat{z}_{i, 16})$) , the surface ($ r_{i}(\hat{z}_{i, 50})$) and the inside ($ r_{i}(\hat{z}_{i, 84})$),  of the object.}
   \label{fig:loss_description}
\end{figure}

\begin{equation}
 L_i = \sum_{i}\| \hat{t}_{16, i}-\epsilon\|^2 +  \| \hat{t}_{50, i}\|^2 + \| \hat{t}_{84, i} + \epsilon\|^2 ,
\end{equation}
We visualize these points in \cref{fig:loss_description}. We also provide an ablation in the supplementary material on the choice of the percentile values. We note that we choose to use values computed from the distribution and not a fixed constant to better account for the scale as well as uncertainties or different levels of details in one scene.

More importantly,  we also  enforce that the learned indicator function is locally smooth and has near constant derivatives close to the zero level set. Namely, the approximated TSDF function should have normals of constant norm  $n_c$ around the median depth. We recall that we are only approximating the behaviour of a SDF in order to run Marching Cubes. We enforce approximated signed distance values at positions computed from the ray distribution. However, the distribution can vary between scenes, ray direction or even regions of the same scenes. Therefore, we do not constrain the function to simulate exact euclidean distances.

\begin{equation}
 L_n = \sum_{i}  \| \| N( r_{i}( \delta))  \| - \mathbf{n}_c\| ^2 ,
\end{equation}
where $N: \mathbb{R}^3 \rightarrow \mathbb{S}^2$ is a function computing the normal at a given point $\mathbf{x}$.
$\delta$ is uniformly sampled value in $[\hat{z}_{i, 16},\hat{z}_{i, 84}]$.
To increase the speed of our method, we compute the normals using finite difference of the SDF network at nearby 3D positions. While other methods such as \cite{verbin2022ref} often compute normals using gradients from the network, their method is significantly more time consuming. 

\paragraph{Normal regularization.} To ensure a smooth surface,  we use the normal regularization used in  RefNeRF \cite{verbin2022ref} on the TSDF normals.
We use the normal smoothness loss:

 \begin{equation}
     L_s = \sum_{i} \| N(r_{i}(\hat{z}_{i, 50}))- \mathbf{\hat{n}}_{50, i} \|^2
 \end{equation}

 Enforcing that the computed normals are close to the estimated normals  produces smoother computed normals as the signals produced by the network are limited in frequency.
 We also use the normal orientation loss:
 
 \begin{equation}
 L_o = \sum_{i} max(0,  N(r_{i}(\hat{z}_{i, 50}))   \cdot \mathbf{d_i}).
 \end{equation} 
 
\paragraph{Appearance.} We train a small appearance network that takes as input shading features $\mathbf{\hat{f}}_i$ and predicted normals $\mathbf{\hat{n}}_i$ predicted by SSAN as well as input view directions $\mathbf{d}_i$ to produce the surface color at a given coordinate from a given view similar to SNeRG \cite{hedman2021snerg}. We supervise this network with the ground truth colors from the training images. We notice that, the projected median depth can be an inaccurate estimation of the surface zero-level set, especially when the angle between the view direction and the surface is small. Therefore, we supervise the color at points projected on the zero-level set. We compute the projection using $4$ steps of gradient descent along the viewing ray using the SDF approximated values, see the supplementary materials for more details. We refer to the depth of  points $r_{i}(\hat{z}_{i, 50})$  projected on the zero level set as $l$.

 \begin{equation}
     L_{color} =\sum_{i} \| \eta(\mathbf{\hat{f}}_{l, i}, \mathbf{\hat{n}}_{l,i },\mathbf{d}_i) - \mathbf{c}_i \|^2,
 \end{equation}
 where $c_i$ is the ground truth pixel color.
 
  The different losses  are combined as a weighted average during training into a final loss $L$.

 \begin{equation}
     L =\alpha L_i + \beta L_n + \gamma L_s + \delta L_o
 \end{equation}

 \paragraph{Training stage.} We explore training the SSAN module in two possible ways. The training process mostly relies on having available depth maps from NeRF. Depths maps can be computed directly during training of the NeRF based method or they can also be rendered from a pretrained NeRF based model. We choose to train our method from the rendered depth percentiles of a pretrained NeRF model. We provide more information about  simultaneously training in the supplementary material. 

 \subsection{Mesh extraction and Real-time rendering}\label{sec-rendering}
 
 \paragraph{Mesh extraction.} The SSAN module converts the radiance field representation of the scene obtained from NeRF to distance field (TSDF) representation. Given the TSDF representation, we can easily reconstruct the surface by using a surface reconstruction algorithm \cite{lorensen1987marching,ju2002dual}. In this paper we use the PyMCubes\footnote{\href{https://github.com/pmneila/PyMCubes}{https://github.com/pmneila/PyMCubes}} implementation of marching cubes.
 Moreover we use either vertex features, for instance on objects from the Blender Synthetic dataset or  build a texture  using per face parametrization for unbounded scenes from \cite{barron2022mipnerf360}. We fill the texture image with appearance features sampled from the SSAN appearance features  at the interpolated face locations.

\paragraph{Rendering.} The triangle mesh geometry extracted from the SSAN can be encoded in any common format like OBJ, glTF, and others. Thus output meshes from out pipeline can be directly integrated in traditional graphics pipeline. The neural view-dependent appearance is added by rasterizing the precomputed texture and feeding the results to the appearance network that produces the final RGB color values. This strategy ensures that the network only performs $w\cdot h$ evaluations and thus can run at a high frame rate on commodity hardware. We measure average FPS of 25 on a workstation and 30 on a MacBook on the Blender Synthetic dataset objects. This approach is similar to SNeRG \cite{hedman2021snerg}, although our method only samples features at the surface boundary instead of aggregating them in a volumetric fashion, similar to MobileNeRF \cite{chen2022mobilenerf}. Contrary to these two methods however, our underlying geometry is by design significantly closer to the ground truth.

\subsection{Implementation details}

We use a backbone similar to Instant NGP \cite{mueller2022instant} for the SSAN module. We divide the module in two separate  branches with separate weights: i) the geometry branch that outputs TSDF approximation and the normal prediction. ii) The appearance branch which outputs color features. Each branch is a separate network with the same architecture.  We use Instant NGP \cite{mueller2022instant} architectures with hash table size $ 2^{19}$, coarsest resolution of 16,  highest resolution of 2048, 15 levels and a number of feature dimension per entry of 2. For the appearance network we use a network of 4 layers of MLP each with width of 32.  We use the JAX framework for our implementation.  We can train and extract a mesh end-to-end in less than an hour using 8 V100 NVIDIA GPUs.  Finally we experimentally set the  hyper-parameters  $n_c=10$ and  $\epsilon = 0.1$.

\section{Evaluation}

We  validate the effectiveness of our approach on synthetic blender scenes from~\cite{mildenhall2020nerf} and real unbounded scenes from~\cite{barron2022mipnerf360}. Our method can be easily integrated to NeRF pipelines and can help with regularizing the underlying 3D geometry to improve its accuracy. To demonstrate this we use the NeRF backbones of Mip-NeRF 360~\cite{barron2022mipnerf360} that we combine with Neural Graphics Primitives ~\cite{mueller2022instant} to speed up  training, and Ref-NeRF~\cite{verbin2022ref}. 
In addition to the results presented in this section, please see the supplementary materials for additional video results and demos.


\subsection{Synthetic blender scenes}


\begin{figure}
  \centering
   \includegraphics[width=1.\linewidth]{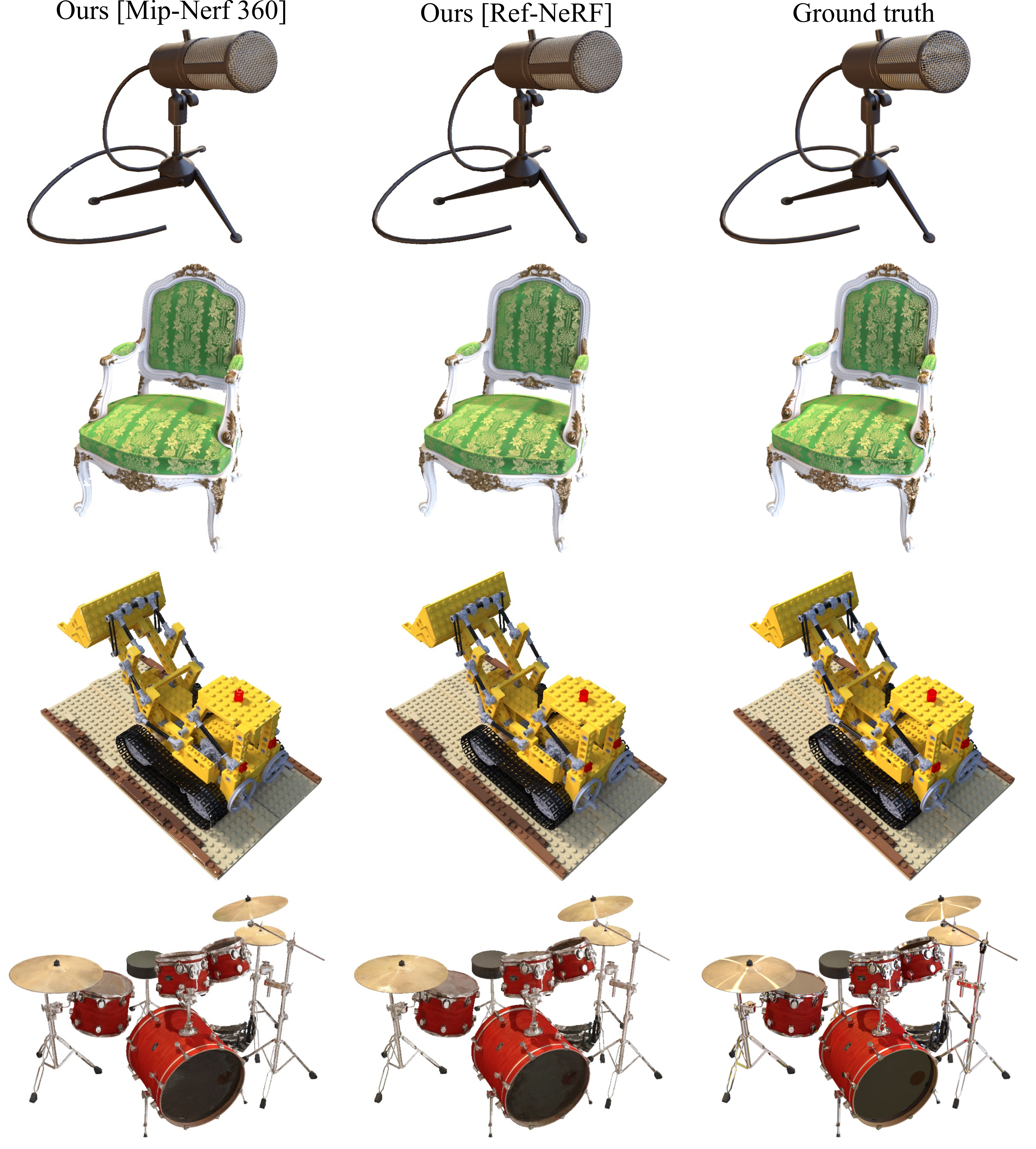}
   \caption{Rendering of the 3D meshes from test views on the Synthetic Blender dataset.}
   \label{fig:blender_qualitative}
\end{figure}

\begin{figure*}
  \centering
    \includegraphics[width=.85\linewidth]{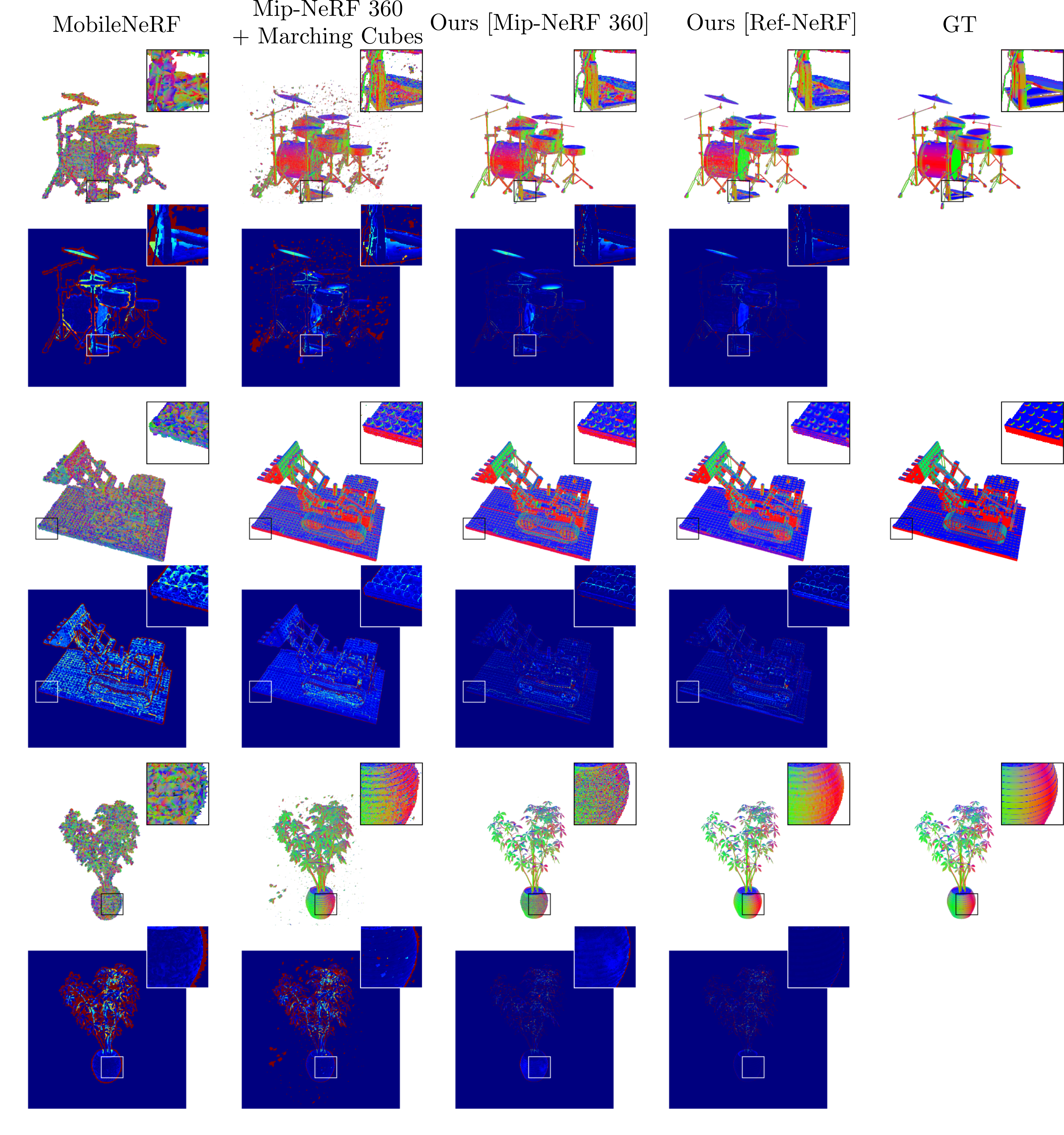}
   \caption{Geometry comparison on the Synthetic Blender scenes~\cite{mildenhall2020nerf}. We visualize mesh normals and the depth absolute difference with the ground truth. We notice that meshes obtained using our method have smoother, more accurate and realistic geometry.}
   \label{fig:blender_geometry_comparison}
\end{figure*}

In this section we focus on the Synthetic Blender dataset from~\cite{mildenhall2020nerf}. We compare our method to state of the art baselines that focus on real time rendering of NeRF: SNeRG \cite{hedman2021snerg}, and MobileNeRF \cite{chen2022mobilenerf}. Additionally, we compare to Mip-NeRF 360~\cite{barron2022mipnerf360} from which we reconstruct the geometry with Marching Cubes (MC) at the same grid resolution as ours and using $0.5$ as the zero-level set in the density grid. Similarly to our method we use Mip-NeRF 360 accelerated with Instant-NGP \cite{mueller2022instant}. For our method, we use vertex based feature representation and a grid of size 1024 inside the bounding box obtained with the ground truth mesh for marching cubes. For the \textit{drums} and \textit{ficus}, we use a higher resolution of 2048 since these two scenes are more detailed. We compare the geometry and appearance produced by each method. We train our method together with the Mip-Nerf 360 backbone and  Ref-NeRF backbone.   

\paragraph{Evaluation metrics.} We measure Chamfer Distance (CD) that we compute using an observability mask similar to \cite{jensen2014large}, as we want to evaluate only in observable regions to for fair comparisons. We construct an observability grid of resolution 256 constructed with rays from training views inside the mesh bounding box. Additionally, we compute the normal consistency (NC) as in \cite{Mescheder_2019_CVPR} as the  absolute value of the dot product between normals from a  point cloud sampled on the ground truth mesh  and the predicted mesh normals. Finally, we also evaluate PSNR to measure  appearance quality.

\paragraph{Comparison.} We present quantitative results on the Blender synthetic dataset in \cref{tab:quantitative_results_blender} and qualitative results in \cref{fig:blender_geometry_comparison}. We observe that our method tends to produce better geometric results in terms of Chamfer Distance and normal consistency. In \cref{fig:blender_geometry_comparison} we show the rendering of normal maps from different meshes (row 1, 3, 5) and the absolute difference with  the ground truth depth rendered from the meshes (row 2, 4, 6). We notice that MobileNeRF produces a \emph{"triangle soup"} that does not accurately reflect the geometry of the object. Computed meshes with marching cubes from pretrained NeRFs lead to more floaters and a surface that is not well aligned with the ground truth as we can notice a red region at the boundary of the depth maps as well as lighter color on the surface.  Moreover, while we are expecting lower appearance quality than other methods, due to the constraints of producing accurate geometry and having real time rendering, we note that rendering quality remains very high and visibly close to the ground truth, as can be seen in \cref{tab:quantitative_results_blender} and \cref{fig:blender_qualitative}. 

\begin{figure}
  \centering
   \includegraphics[width=1.\linewidth]{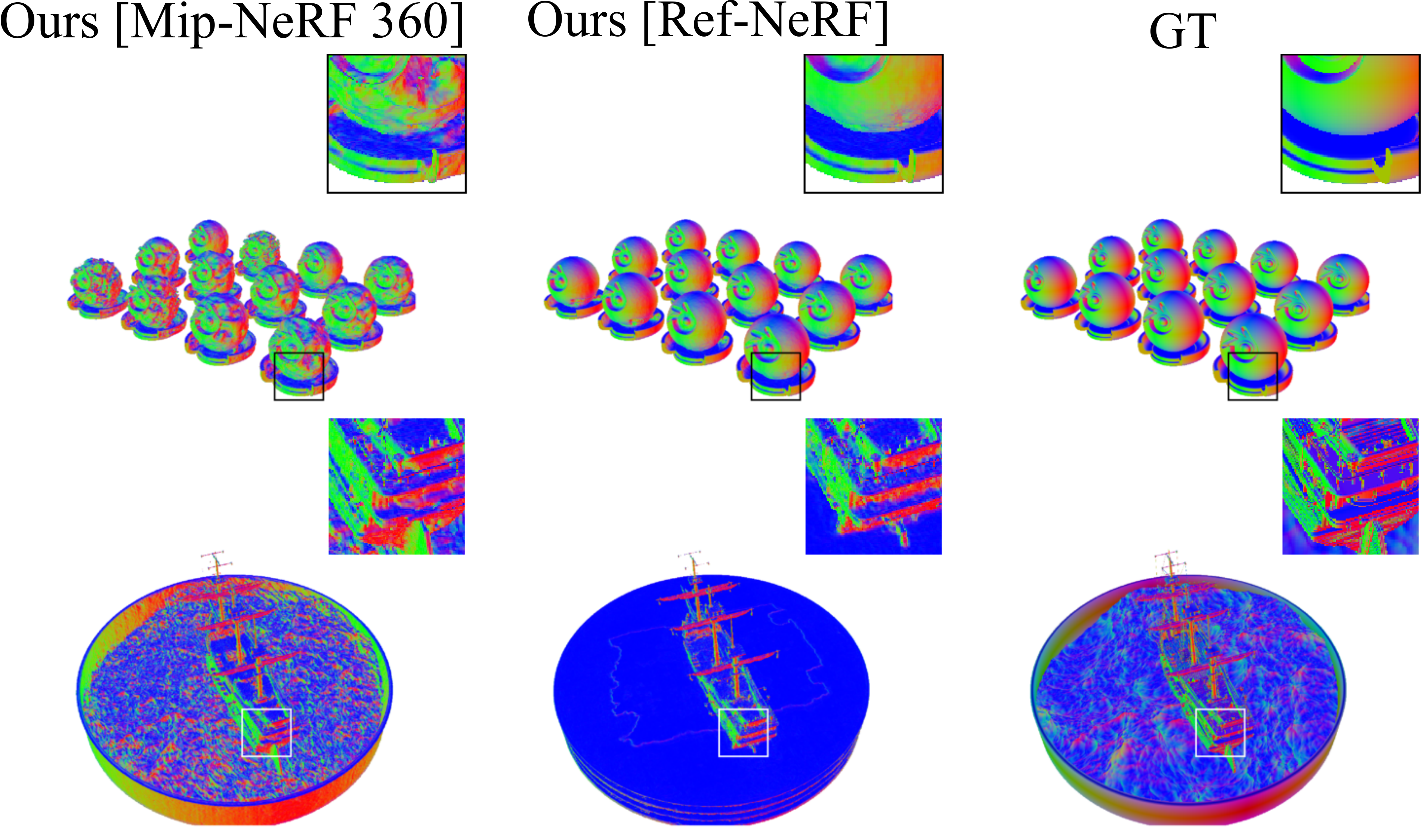}
   \caption{Effect of the NeRF backbone.}
   \label{fig:blender_backbone}
\end{figure}

\paragraph{Effect of the backbone} We observe that using the appropriate NeRF based backbone can improve the quality of the final mesh significantly in \cref{fig:blender_backbone}. In particular, shiny objects such as the  Materials scene  highly benefit from the Ref-NeRF backbone as the method is specifically designed to deal with such objects (bottom of \cref{fig:blender_backbone}). On the other hand, the Ref-NeRF backbone can sometimes fail to approximate the right geometry due to the powerful appearance network while Mip-NeRF 360 produces more realistic geometry, for instance on the ship scene (top  of \cref{fig:blender_backbone}). This effect is also visible in \cref{tab:quantitative_results_blender} as the Mip-NeRF 360 backbone produces more accurate geometry for the Ship scene and Ref-NeRF backbone produces more accurate geometry for the Materials scene. Finally we also observe the effect of the backbones in \cref{fig:blender_geometry_comparison} as the ficus and drums scenes have shiny features that are better reconstructed when using the Ref-NeRF backbone.

\begin{table*}[t!]
    \centering
    \resizebox{\linewidth}{!}{
    \begin{tabular}{cc|cccccccccc}
        Metric & Method & Chair & Drums & Ficus & Hotdog & Lego & Mats. & Mic & Ship & Avg  \\
         \hline
      
          \multirow{4}{*}{PSNR $\uparrow$} & MobileNeRF \cite{chen2022mobilenerf} & \textbf{34.09} & \textbf{25.02} & \textbf{30.20} & \textbf{35.46} & \textbf{34.18} & 26.72 & 32.48 & \textbf{29.06} & \textbf{30.90} \\
          & SNeRG \cite{hedman2021snerg}& 33.24 & 24.57 & 29.32 & 34.33 & 33.82 & \textbf{27.21} & \textbf{32.60} & 27.97  & 30.38\\

          &\textbf{Ours}[Mip-Nerf 360] &31.44& 23.42& 26.46& 31.63& 29.01& 20.36& 27.41& 26.41 & 27.02\\
      
         &  \textbf{Ours} [Ref-NeRF] &31.93 & 23.49& 25.95& 32.38 &28.89 &23.30 &27.83& 24.70 & 27.31\\
       
        \hline
         \multirow{4}{*}{CD $\downarrow$} &Mip-Nerf 360 + Marching Cubes&0.016&0.047&0.032&0.027&0.024&0.019&0.022&0.044&0.029\\
         &MobileNeRF & 0.015&0.02&0.018&\textbf{0.019}&0.020&0.017&0.022&0.029&0.020\\
       
         &  \textbf{Ours} [Mip-Nerf 360] &0.012&0.018&0.012&0.026&0.018&0.023&0.016&\textbf{0.019}&\textbf{0.018}\\

         & \textbf{Ours} [Ref-NeRF] &\textbf{0.010}&\textbf{0.018}&\textbf{0.008}&0.042&\textbf{0.015}&\textbf{0.014}&\textbf{0.013}&0.104&0.028\\
        \hline
        
         \multirow{4}{*}{NC $\uparrow$}  &Mip-Nerf 360 + Marching Cubes&0.808&0.750&0.779&0.815&0.645&0.836&0.763&0.696&0.762 \\
         &MobileNeRF&0.615&0.550&0.655&0.605&0.546&0.49&0.519&0.535&0.565\\
       
          &  \textbf{Ours} [Mip-Nerf 360]&0.855&0.793&0.809&0.843&0.711&0.824&0.762&\textbf{0.772}&0.796\\

         &\textbf{Ours} [Ref-NeRF] & \textbf{0.857}&\textbf{0.803}&\textbf{0.865}&0.854&\textbf{0.714}&\textbf{0.908}&\textbf{0.773}&0.611&\textbf{0.798}\\
    \end{tabular}}
    \caption{Quantitative results on the Blender Synthetic dataset \cite{martin2021nerf}. We measure appearance with PSNR and geometry with Chamfer Distance (CD) and normal consistency (NC). Our method achieves better geometric reconstruction overall while producing reasonable PSNR.}
    \label{tab:quantitative_results_blender}
\end{table*}

\subsection{Real unbounded scenes}

\begin{table*}[t!]
    \centering
    \begin{tabular}{cc|ccccccc}
        Metric & Method& Bicycle & Garden& Stump& Counter & Room & Bonsai & Kitchen \\
         \hline
         \multirow{4}{*}{PSNR $\uparrow$} & NeRF \cite{mildenhall2020nerf} & 21.76 & 23.11 & 21.73 & 25.67 & 28.56 & 26.81 & 26.31\\
          & MobileNeRF \cite{chen2022mobilenerf} &  21.70  &23.54 &23.95 & - & - & - & -\\
         & Ours & 21.15 &  22.91 & 22.66 &  20.00 &  26.13 &  25.58 & 23.59 \\
    \end{tabular}
    \caption{Novel view evaluation on real unbounded scenes from~\cite{barron2022mipnerf360}.}
    \label{tab:quantitative_results_unbounded}
\end{table*}

\begin{figure}
  \centering
   \includegraphics[width=0.80\linewidth]{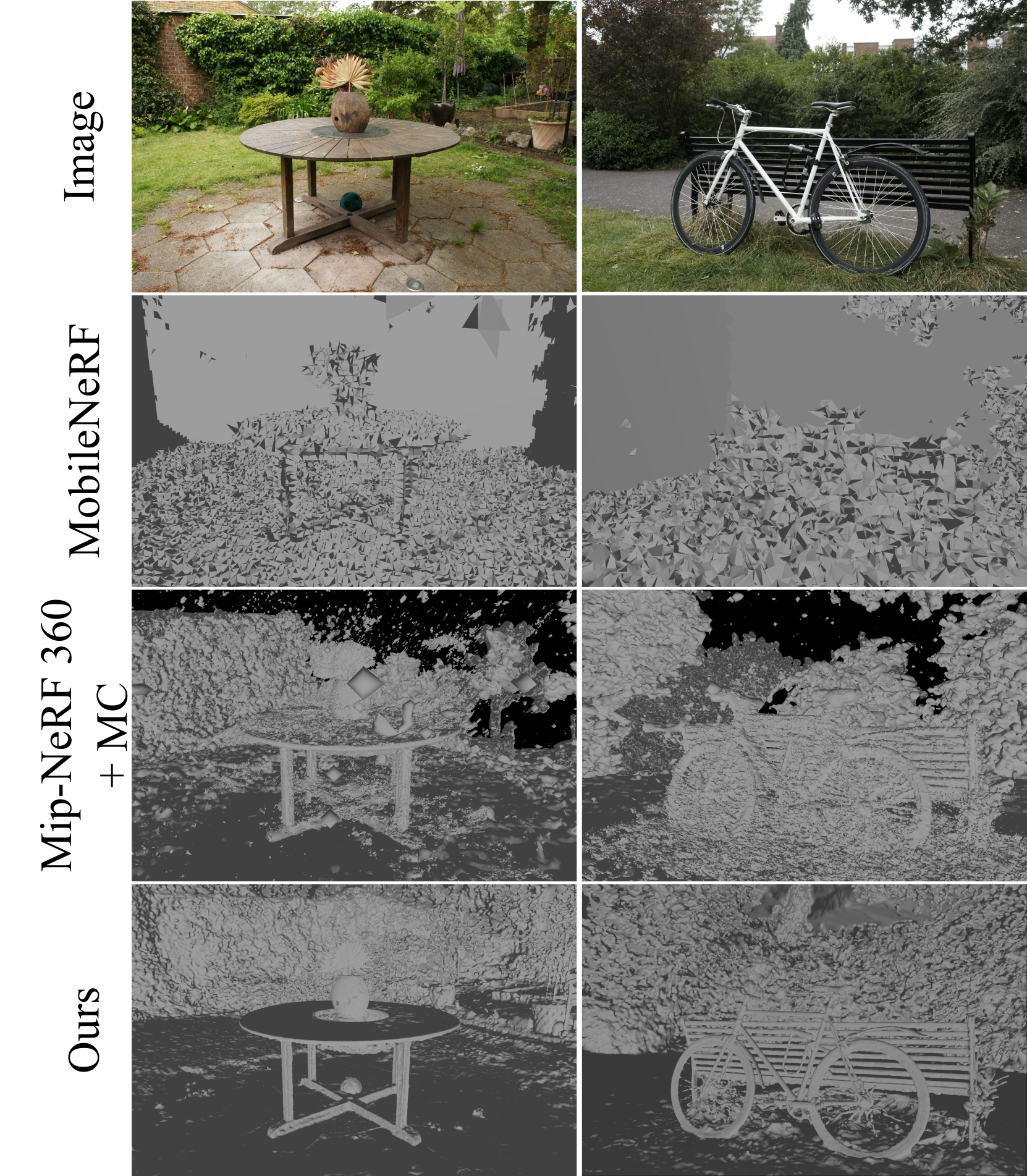}
   \caption{Results on scenes from the unbounded dataset \cite{barron2022mipnerf360} shown on the left. Rendered color images from our 3D meshes are shown in the right. Note that meshes from our method captures the scene geometry much better compared to MobileNeRF~\cite{chen2022mobilenerf} and \cite{barron2022mipnerf360}.}
   \label{fig:geometry_rendering}
\end{figure}

\begin{figure}
  \centering
   \includegraphics[width=0.8\linewidth]{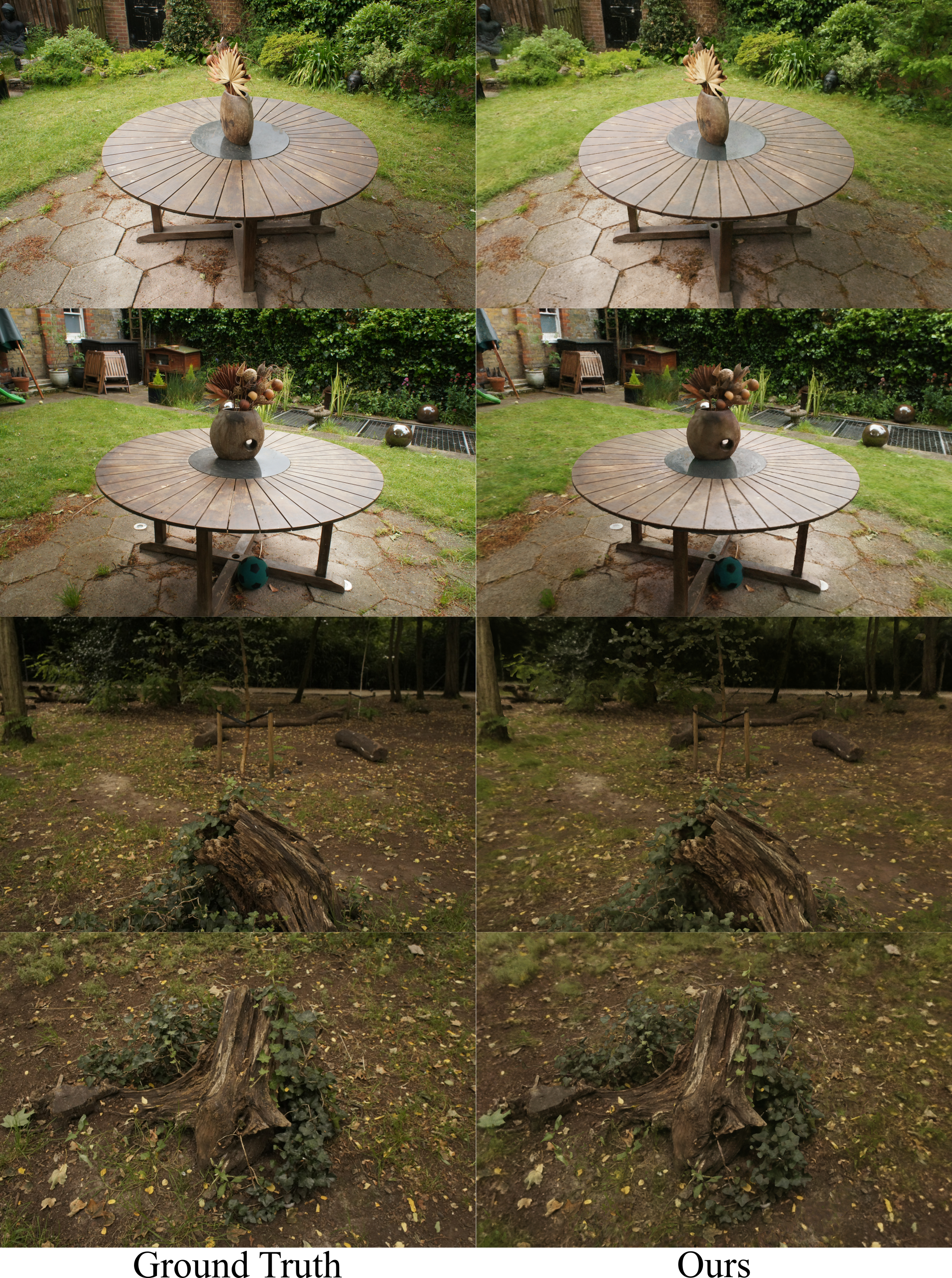}

   \caption{Rendered novel views of our method on real unbounded scenes from Mip-NeRF 360~\cite{barron2022mipnerf360}. Note that rendered images are of high fidelity and captures view dependent effects while being rendered real-time on commodity hardware.}
   \label{fig:unbounded_qualitative}
\end{figure}

In this section we show results on unbounded scenes from Mip-NeRF 360 \cite{barron2022mipnerf360}. We evaluate on the publicly available scenes. We use Mip-NeRF 360 \cite{barron2022mipnerf360} accelerated with NGP from~\cite{mueller2022instant} as our NeRF backbone architecture for experiments on this dataset. Note that there is no ground truth geometry available for these scenes so we do not evaluate the geometry numerically.

When evaluating on unbounded scenes, we extract the mesh using marching cubes separately for both the scene foreground and scene background. To limit the total size of the mesh, we use a foreground box of resolution 1024 to mesh the foreground and separately use a background box of resolution 2048 to mesh the background. Due to the higher number of vertices we build a feature texture to retain image resolution. We construct a per face parametrization of the texture and map each face to a right-angled triangle of side 4 pixels in the texture image, we provide more details in the supplementary materials. 

 \cref{fig:geometry_rendering} provides a qualitative comparison of the underlying geometry obtained from our method with other methods. As shown in \cref{fig:geometry_rendering}, our method produces more accurate geometry compared to MobileNeRF meshes and Mip-Nerf 360 meshes reconstructed via marching cubes. A key contributor of this advantage is the explicit regularization of the SDF via our SSAN. Rendered novel views from our method are shown in \cref{fig:unbounded_qualitative}. We observe in  \cref{fig:unbounded_qualitative} and \cref{tab:quantitative_results_unbounded} that our method remains of good visual quality even though it is constrained by accurate geometry. 

\subsection{Physics-based applications}
Since our method produces accurate 3D meshes, several applications such as scene editing and physics simulation can be easily conducted with traditional graphics and simulation pipelines. In \cref{fig:teaser}, we show simple scene editing where we combine meshes extracted from \textit{Chair}, \textit{Hotdog}, \textit{Lego} (Blender dataset), \textit{Bonsai} and \textit{Garden} (Mip-Nerf 360 dataset) sequences using our method.  We further show a simulation of a cloth falling on objects from Blender Synthetic dataset. Please see the supplementary material for additional video results and demos.

\section{Conclusion}

In this work we propose a novel approach to extract geometrically accurate meshes from NeRF based architectures. Our SSAN model can be trained from any NeRF architecture  without a significant penalty in training time. Our neural mesh representation can be rendered  at high frame rates on commodity hardware. Thanks to their geometric accuracy, our extracted meshes can be quickly   visualized, and also be used in physically accurate settings in simulations, to compute accurate occlusions and interactions with other objects.
Our work is nevertheless limited in some aspects: rendering detailed surfaces causes us to generate meshes with a high number of faces and vertices, thus, our method would gain from using an adaptative mesh reconstruction strategy. Moreover, on large (unbounded) and detailed scenes, we are limited to lower resolutions to not hinder  the overall size of the model. Finally, the appearance network that we are using is a non-standard component that requires customization of existing renderers. Being able to learn the underlying materials, as methods like~\cite{boss2022-samurai} would further simplify the integration of meshes in other tools.



{\small
\bibliographystyle{ieee_fullname}
\bibliography{egbib}
}

\appendix
{\begin{flushleft}\LARGE \textbf{Appendix} \end{flushleft}}

\begin{table*}
\begin{center}
\begin{tabular}{l|l|cccccccccc|}
\hline
 Metric & Method & Chair & Drums & Ficus & Hotdog & Lego & Mats. & Mic & Ship & Avg  \\
 \hline
 PSNR &\textbf{Ours}[post-processing] &31.44& 23.42& 26.46& 31.63& 29.01& 20.36& 27.41& 26.41 &27.02 \\
 &  \textbf{Ours} [joint-training]&\textbf{31.54}&\textbf{23.83}&\textbf{26.70}&\textbf{32.16}&\textbf{29.21}&\textbf{21.54}&\textbf{27.89}&\textbf{26.63} & \textbf{27.44}\\
 \hline
 CD &  \textbf{Ours} [post-processing] &\textbf{0.012}&\textbf{0.018}&\textbf{0.012}&\textbf{0.026}&0.018&0.023&0.016&0.019&0.018\\
 &  \textbf{Ours}[joint-training]  &0.013&\textbf{0.018}&0.013&0.030&\textbf{0.017}&\textbf{0.021}&\textbf{0.015}&\textbf{0.028}&\textbf{0.019}\\
   \hline
  
  NC &  \textbf{Ours} [post-processing]&\textbf{0.855}&0.793&0.809&0.843&\textbf{0.711}&\textbf{0.824}&\textbf{0.762}&\textbf{0.772}&0.796\\
&  \textbf{Ours} [joint-training] &0.842&\textbf{0.803}&\textbf{0.844}&\textbf{0.860}&0.697&0.809&0.758&0.769&\textbf{0.798}\\
 \hline
\end{tabular}
\end{center}
\caption{Effect of training stages on the blender synthetic dataset.}
\label{tab:jointtraining}
\end{table*}

\begin{table}
\begin{center}
\begin{tabular}{l|c|c|c}
\hline
 & PSNR &CD & NC \\
\hline
 w/o appearance network separation& 26.74 &\textbf{0.018} & \textbf{0.802} \\
 w/ percentiles 5 \& 95& 27.01 &\textbf{0.018} & 0.796 \\
 w/ percentiles 25 \& 75& \textbf{27.02} &\textbf{0.018} & 0.796 \\
 w/o projection & 26.82& \textbf{0.018}& 0.798\\
 Ours& \textbf{27.02} &\textbf{0.018} & 0.796 \\
\hline
\end{tabular}
\end{center}
\caption{Ablation study on the blender synthetic dataset.}
\label{tab:ablation_percentiles}
\end{table}

\section{Joint training}  \label{sec:jointtraining}
 
 We choose to train our method from a pre-trained NeRF method. In this section we measure the performance of training our method simultaneously to an un-trained Mip-NeRF 360 using instant NGP~\cite{mueller2022instant}.  We observe in  \cref{tab:jointtraining} that joint training leads to better appearance. However, due to the increased simplicity we choose to evaluate our method as post-processing from a pre-trained NeRF method.

\section{Ablation study}   \label{sec:ablation}

We evaluate the effect of some of our design choices for our method trained with a Mip-NeRF 360 with Instant NGP on the Blender Synthetic dataset in\cref{tab:ablation_percentiles}.  
\paragraph{Effect of choice of percentiles.} We train our method on the blender synthetic dataset with different choices of percentile values. We notice that this choice only has a minor effect (\textit{c.f.}~\cref{tab:ablation_percentiles} row 2, 3, 5). 

\paragraph{Effect of separate appearance network.} We train our method with a single network to represent both the geometry and the appearance. We notice that separating the network leads to a small improvement over PSNR whilst geometric metrics remain comparable  \cref{tab:ablation_percentiles} row 1, 5. 

\paragraph{Effect of  the projection.} We train our method without the projection defined in Equation 10. We notice that it leads to a small drop in PSNR while geometric metrics are not much affected (\textit{c.f.}~\cref{tab:ablation_percentiles} row 4, 5).

\section{Additional implementation details} \label{sec:implementationdetails}
\subsection{Projection on the zero level set }
The median depth computed from a pretrained NeRF can sometimes be inaccurate, especially when the angle between the view direction and the surface is small. To get a better approximation of the zero level-set of the surface we apply 4 steps of  gradient descent along the viewing ray using the TSDF predicted value. We define one step applied to a 3D point $p$ as :

\begin{equation}
p' = p + \alpha * \hat{t},
\text{ with }
[\hat{t}, \mathbf{\hat{n}}, \mathbf{\hat{f}}] = \phi(\mathbf{p}),
\end{equation}
where $\alpha$ is a small constant. 
\subsection{Per face parametrization}
\begin{figure}
  \centering
   \includegraphics[width=0.3\linewidth]{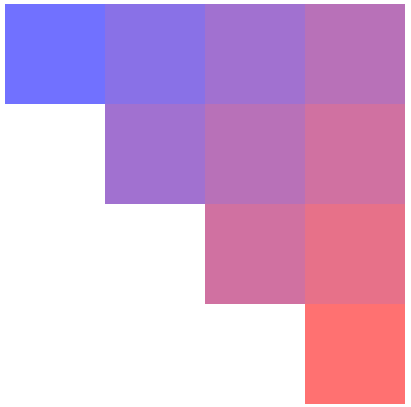}
   \caption{Representation of each face of the  mesh in the texture image. Each face is mapped to a right angled triangle containing features predicted by SSAN.}
   \label{fig:parametrization}
\end{figure}

When evaluating on unbounded scenes from \cite{barron2022mipnerf360}, we represent scene features using textures with per face parametrization. We map each face in the triangle mesh to a right angled triangle of side 4 in the texture image. We show the face shape in \cref{fig:parametrization}. Each pixel feature  $\mathbf{\hat{f}}$ is computed using the SSAN at the corresponding 3D position.  

\subsection{Training times}
Our method has 4.8 minutes training time on average on the Blender Synthetic Dataset.

\end{document}